\pdfoutput=1
\documentclass[11pt]{article}
\usepackage[preprint]{acl} % "final" for camera ready, "review" for review, "preprint" non-anonymous.
\usepackage{times}
\usepackage{latexsym}
\usepackage[T1]{fontenc}
\usepackage[utf8]{inputenc}
\usepackage{microtype}
\usepackage{inconsolata}
\usepackage{graphicx}
\usepackage{algorithm}
\usepackage{algpseudocode}
\usepackage{multirow}
\usepackage{amsmath, amsfonts, amssymb}
\usepackage{fancyhdr}
\title{Language Models are Crossword Solvers}

\author{
    \textbf{Soumadeep Saha}, \textbf{Sutanoya Chakraborty}, \textbf{Saptarshi Saha}, \textbf{Utpal Garain}
    \\
    \\
  Indian Statistical Institute, Kolkata
\\
  \small{\textbf{Correspondence:} \href{mailto:soumadeep.saha97@gmail.com}{soumadeep.saha97@gmail.com}}
}

\begin{document}
\maketitle

% Changes:
% Capitalization of large language models.
% Removed mention of SoTA from line 8-13 as per reviewer suggestion.
\begin{abstract}
    Crosswords are a form of word puzzle that require a solver to demonstrate a 
    high degree of proficiency in natural language understanding, wordplay, 
    reasoning, and world knowledge, along with adherence to character and length 
    constraints. In this paper we tackle the challenge of solving crosswords with 
    large language models (LLMs). We demonstrate that the current generation of 
    language models shows significant competence at deciphering cryptic 
    crossword clues and outperforms previously reported state-of-the-art (SoTA) 
    results by a factor of 2-3 in relevant benchmarks. We also develop a search 
    algorithm that builds off this performance to tackle the problem of solving 
    full crossword grids with out-of-the-box LLMs for the very first time, 
    achieving an accuracy of 93\% on New York Times crossword puzzles. 
    Additionally, we demonstrate that LLMs generalize well and are capable of 
    supporting answers with sound rationale.
\end{abstract}

\section{Introduction}
\label{sec:intro}

\begin{figure*}
    \begin{center}
        \includegraphics[width=0.98\textwidth]{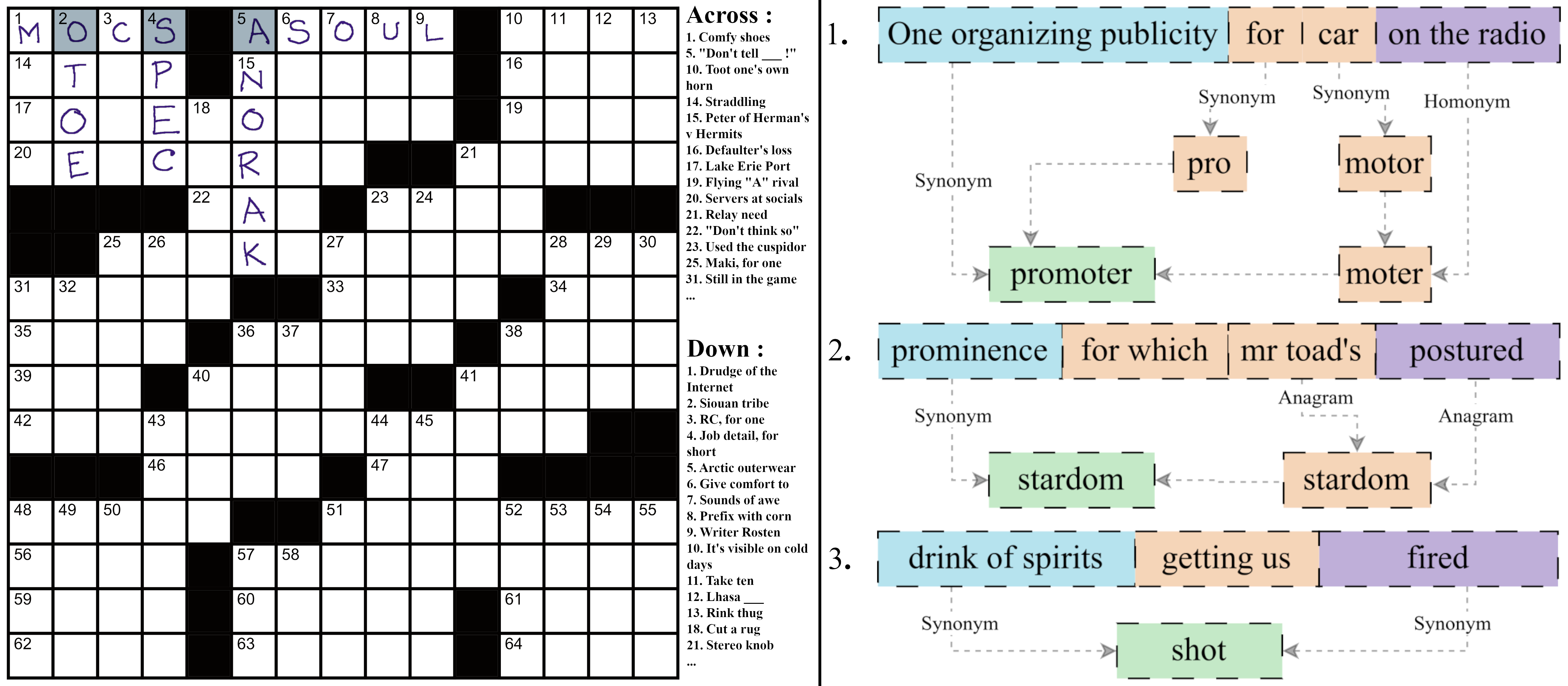}
    \end{center}
    \caption{
        \textbf{Example of a crossword puzzle (left) and cryptic clues (right).}
        (\emph{left}) The grid must be filled up with answers from the semantic clues 
        provided. The gray highlighted squares produce additional constraints,
        e.g., first character of the answer to clue 1 (across) and clue 4 (down)
        must be the same. Example by Fred Piscop.
        (\emph{right}) In cryptic crosswords, the clues involve some form of wordplay 
        and synonyms and often involve world knowledge. Examples are taken
        from the cryptonite dataset (see Appendix \ref{sec:appendix-hev} for more).
    }
    \label{fig:crossword_subtask}
\end{figure*}

\footnote{Code, data, etc., can be found at 
\url{https://www.github.com/espressovi/LMCrossword}}Crossword puzzles are a type
of word game that typically take the form of a square grid of white and black
boxes. The objective of the puzzle is to fill the white boxes with letters from
words or phrases based on the provided clues (see Figure 
\ref{fig:crossword_subtask}). Although crosswords come in a variety of styles, 
the two most popular ones are the American style, or \emph{straight} crosswords, 
and \emph{cryptic} crosswords.

Modern LLMs demonstrate astounding skills in reasoning, coding, wordplay, 
question answering (QA), and a multitude of other tasks \citep{wei2022emergent,
survey}. Despite the plethora of applications for LLMs seen today, their ability
to generate language in a constrained setting remains relatively uncharted, and
the ability to direct the generation process in order to meet certain criteria
remains a challenge \citep{poemgeneration}. Consider the task of poem generation
as an example, where, in addition to thematic aptness, constraints on rhyme or
meter must be adhered to. Further, for certain kinds of poems like haikus,
sonnets, or even song lyrics, restrictions on length, syllable counts, or
patterns of stressed/unstressed syllables apply. Constraints also arise when
dealing with formal languages \citep{automatabased}, which is increasingly
relevant in the LLM zeitgeist given their use in coding or interacting with
databases, interpreters, APIs, etc. With a growing body of literature studying
LLMs as agents \citep{mineagent, embagent}, language generation must follow
environmental or physical constraints. For wider proliferation of LLM
applications, they must demonstrate the ability to adhere to such
domain-specific constraints (e.g., constraints imposed by knowledge graphs,
tabular data, action spaces, etc.).

Solving \emph{crosswords} requires proficiency in understanding contextual clues, 
semantics, wordplay, character manipulation, arithmetic, and reasoning (see Figure 
\ref{fig:crossword_subtask} \emph{(right)}), along with satisfying constraints 
such as length limitations and character overlaps (see Figure \ref{fig:crossword_subtask}).
Given the multi-faceted nature of this task, it can serve as a testbed for
studying constrained language generation. Since solving crosswords requires
proficiency in several desirable areas, and identifying shortcomings can readily
benefit other linguistic applications, in this paper we attempt to analyze LLMs'
ability to solve crossword puzzles, with the \textbf{primary goal of
understanding strengths and weaknesses demonstrated by SoTA LLMs}. Our
contributions in this study are as follows:

$\diamond$ \textbf{We perform extensive analysis to understand how well LLMs
can answer crossword clues} based on the provided contextual information and 
constraints. We show that current-generation SoTA LLMs demonstrate massively 
improved performance when compared to previous SoTA baselines, \emph{
outperforming some previous benchmarks by a factor of 2 or 3, without any 
fine-tuning}.

$\diamond$ \textbf{Our analysis shows that the ability of LLMs to adhere to
length constraints, which is a critical component of solving crossword puzzles, 
is somewhat limited.} We present results demonstrating that an LLM's ability to 
count the number of constituent characters in a word degrades with the 
prevalence of the word. This suggests that their ability to perform this task
does not generalize well. Additionally, we show that this is a significant 
limiting factor for crossword solving, since even when the generated answers are 
semantically similar, the length constraint is frequently not obeyed.

$\diamond$ \textbf{We devise a simple algorithm to tackle the full problem of 
solving crosswords with an LLM-guided search and incorporating grid information,}
solving straight crossword puzzles with an accuracy of \textbf{93.1\%}. This algorithm 
exploits the constraints imposed by previously generated answers to improve future 
answers and achieves results that are not far behind those of specialized 
automated crossword-solving systems. This algorithm also improves the 
clue-deciphering capability of LLMs, more than doubling the baseline performance.

$\diamond$ \textbf{We perform human evaluation and further experiments with 
post knowledge-cutoff datasets} to assess the soundness of logical reasoning, 
potential pitfalls, and generalizability. We find that SoTA LLMs appear to 
generalize well and provide sound logical reasoning with 74\% of correct answers.

We believe our work has been a major step towards demonstrating the power of LLMs 
with regard to crossword solving, and with future advances in certain key areas, 
LLMs' performance in this task will be comparable to, if not better than, human 
experts and specialized systems. Our algorithmic approach of combining an LLM
with a search strategy might be extended to linguistic problems with 
domain-specific constraints.

\section{Background}
\label{sec:background}

Traditional approaches to solving \emph{straight} crossword puzzles involve two
key components: a candidate answer proposal system and a grid-filling algorithm
\citep{webcrow, drfill}. Answer proposal systems typically use similarity search
on large clue-answer databases, fine-tuned language models, or a combination of 
both \citep{proverb1, arsehole}. Grid-filling relies on variants of constraint 
satisfaction problem (CSP) algorithms. For instance, the system \emph{Proverb} 
\citep{proverb} achieves a 98.1\% letter accuracy on New York Times (NYT) 
crosswords. \citet{arsehole} fine-tuned BERT and ByT5 on 6.4 million clue-answer 
pairs, and using a belief propagation algorithm, achieved a 99.7\% letter 
accuracy. \citet{dacross} established benchmarks with foundational language 
models and highlighted this task as ``\emph{... a new high bar for AI systems.
}''. Note that this work \textbf{does not attempt to create an improved 
automated crossword-solving system} but seeks to analyze foundational LLMs' 
ability at this complex language task.

Cryptic crosswords are more formidable and involve extensive wordplay such as 
anagrams, splicing, homophones, and puns. Traditional algorithms with large 
clue datasets and a context-free grammar parser \citep{deits} have shown poor 
performance, achieving only 7\% accuracy \citep{llmcryptic}. Recent studies 
have explored using large language models (LLMs) to solve cryptic crossword clues. 
\citet{cryptonite} created a large dataset of cryptic crossword clues from UK 
dailies,  \emph{The Times} and \emph{The Telegraph}, and fine-tuned a T5-Large 
\citep{t5} model to establish baseline performances. They used a training split 
where answers in the training and test sets were mutually exclusive to prevent 
memorization. \citet{curriculum} curated a dataset from \emph{The Guardian} 
and fine-tuned a T5-Large model using curriculum learning, showing performance 
improvements. They critiqued \citet{cryptonite}'s approach, arguing that a 
disjoint train-test split is insufficient for teaching models to solve cryptic 
crosswords, as models exhibit ``... robustness to plural and other 
inflections.'' Instead, they proposed grouping similar-root words together in a 
split, noting that this more stringent criterion led to reduced performance.

The most recent work by \citet{llmcryptic} presents results with recent LLMs
like Mistral-7B \citep{mistral}, LLaMA2-7B \citep{llama2}, and ChatGPT 
\citep{chatgpt} in few-shot settings and also by fine-tuning the Mistral model.
They report that ChatGPT surpasses other models with an accuracy of $9.5\%$, 
which demonstrates a significant performance gap with respect to human experts, 
who solve $99\%$ of cryptic crossword clues \citep{llmcryptic}. They point out 
key limitations in their work, like the limited set of LLMs used and the
potential for data contamination. 

These recent works on cryptic crossword solving with LLMs highlight a 
\emph{significant performance gap between LLMs and human experts}. However, 
these recent works approach the problem as a question answering (QA) task and 
ignore constraints imposed by the grid. This needs to be investigated further 
since it is yet to be seen whether a suitable approach that integrates 
constraint information into LLMs can yield significant performance benefits.

As for straight crosswords, \citet{dacross} attempted to solve NYT crossword 
puzzles with language models and a Satisfiability Modulo Theory (SMT) solver 
with limited success, and resorted to culling the crossword grid based on 
candidate generations and ground truth answers. We present an algorithm employing
LLM-guided search, which is significantly more successful at this task. With 
advancements in LLM capabilities and methods like ours, we believe LLMs could 
soon outperform humans in solving cryptic crosswords too.

\section{Analyzing Crossword Clue Solving}
\label{sec:analysis}

The first step in solving a crossword puzzle is deciphering its clues, so 
following in the scheme of \citet{curriculum}, \citet{llmcryptic}, 
\citet{dacross}, and \citet{cryptonite}, in this section, we first explore this 
as a QA task. We analyze the performance of several LLMs at different scales 
with variations of this task.

\subsection{Datasets}

Most of our analysis is performed on three crossword puzzle datasets. The first one
covering \emph{straight} or American-style crosswords was curated by us from the 
very popular and long-running puzzle section of the \emph{NYT}. The other two, 
namely, \emph{Cryptonite} by \citet{cryptonite} and \emph{word-init-disjoint} 
(abbreviated as \emph{Init}) by \citet{curriculum}, cover cryptic crossword 
puzzles. Note, the methodological differences between \emph{Cryptonite} and 
\emph{Init}, as discussed in Section \ref{sec:background}, are not pertinent for 
us since we \textbf{do not perform any training}. A bulk of the results are 
reported on the \emph{NYT} dataset we curated for straight crosswords and 
\emph{Init} for cryptic crosswords since it was found to be more challenging 
than \emph{Cryptonite} for LLMs. We drew 2000 randomly chosen samples to report 
results, and in-context examples were also randomly selected from a large pool 
of samples disjoint from the testing set. Further details can be found in 
Appendix \ref{sec:appendix-data}.

\subsection{Answering Crossword Clues}
\label{sec:ansclue}

\begin{figure*}[ht]
    \begin{center}
        \includegraphics[width=0.32\textwidth]{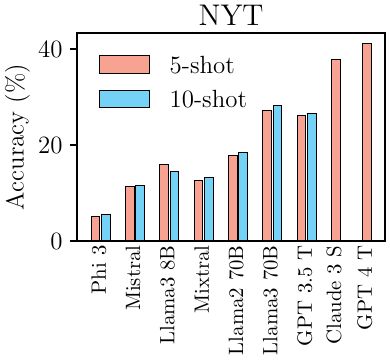}
        \includegraphics[width=0.32\textwidth]{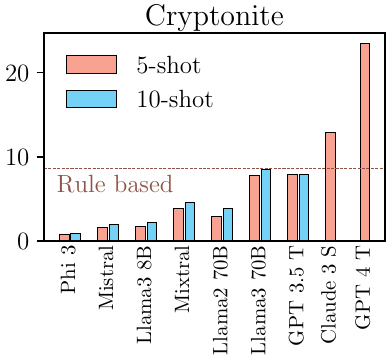}
        \includegraphics[width=0.32\textwidth]{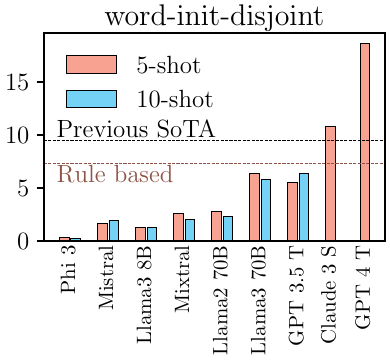}
    \end{center}
    \caption{\textbf{Analyzing LLMs' ability to generate answers from crossword clues.}
    We test LLMs at different scales on the NYT, Cryptonite, and \emph{Init}
    dataset with 5-shot, 10-shot prompts. All results are with \texttt{T=0.5}, 
    further details about experimental protocols can be found in Appendix 
    \ref{sec:appendix-exp}.
    }
    \label{fig:semantic_context}
\end{figure*}

In this experiment we only provided the LM with the query clue and length 
information (alongside instructions and in-context examples), with the expectation 
that it will produce the corresponding answer. We report results with Phi 3 3.8B
Instruct \citep{phi3}, Mistral 7B Instruct \citep{mistral}, Llama 2 70B 
\citep{llama2}, Llama 3 8B Instruct, Llama 3 70B \citep{llama3}, Mixtral 8x7B 
\citep{mixtral}, Claude 3 Sonnet \citep{claude}, GPT 3.5 Turbo, and GPT 4 Turbo 
\citep{gpt4} to cover a wide range of parameter scales and a mix of open-weights
and proprietary models. We investigate the performance of these models with 
few-shot prompts (5-shot, 10-shot)\footnote{Further increases (25-shot) did not 
yield performance benefits.} on samples from the NYT dataset (ours), 
\emph{Cryptonite}, and the \emph{Init} split (see Figure \ref{fig:semantic_context}).

We find that the performance difference between 5-shot and 10-shot\footnote{We 
elected not to perform experiments with 10-shot prompts on Claude and 
GPT-4-Turbo due to budget constraints.} prompted answers is not significant 
(see Figure \ref{fig:semantic_context}). We also note that there is an 
appreciable difference (up to 5\%) in the performance of models across the board 
between \emph{Cryptonite} and \emph{Init}, with LLMs showing diminished 
performance on the \emph{Init} dataset.

The models demonstrate improved performance with scale across datasets and, in 
particular, show remarkable improvement on the NYT dataset with Llama 3 70B, 
GPT 3.5 Turbo, Claude 3 Sonnet, and GPT-4-Turbo achieving 27.2\%, 26.05\%, 37.7\%
and \textbf{41.2\%} accuracy (exact match), respectively. The performance of 
LLMs on cryptic crosswords is worse compared to straight 
crosswords; however, Claude 3 Sonnet and GPT-4-Turbo \textbf{outperform previous 
SoTA results} on both the cryptic crossword datasets, achieving an accuracy of 
12.9\%, \textbf{23.5\%} on Cryptonite and 10.8\%, \textbf{18.7\%} on \emph{Init}
respectively. The performance of GPT-4-Turbo is rather extraordinary, with a 
\textbf{1.97}$\times$ improvement over previous SoTA (9.5\% reported by 
\citet{llmcryptic}, see Figure \ref{fig:semantic_context}). 

\subsection{Exploiting Partially Filled Grids}
\label{sec:partgrid}

In the course of solving a crossword puzzle, we encounter intermediate states
where some of the clues have been deciphered. Crossword solvers exploit these 
clues in order to inform their decisions about future answers. For example,
in Figure \ref{fig:crossword_subtask} \emph{(left)}, when we want to solve for 
the clue in position 14 (across), we can use the characters from position 2 
(down) and 4 (down) to narrow down the set of possible answers to only those 
that fit the template ``\texttt{\_ T \_ P}''. In this section, we study LLMs'
performance at exploiting these constraints to arrive at better answers 
(see Table \ref{tab:hinted_results}).

For this experiment we chose the best performing open-weights and proprietary 
models, i.e., LLaMA 3 70B and GPT-4-Turbo, respectively. We also report results
on some smaller models to see if the trends hold. We report results on the NYT
dataset and \emph{Init} dataset with 5-shot prompts. For each query, $k$\% of 
the characters (letters) of the answer is provided alongside the clue and 
expected length of the answer, and the LLMs are expected to ``unmask'' the 
remaining characters using the provided constraint information and the crossword 
clue\footnote{\texttt{T=0.5, max-tokens=10}, more details in Appendix 
\ref{sec:appendix-exp}.}.

\begin{table*}[ht]
    \begin{center}
        \begin{tabular}[c]{|l|l|l|l|l|l|l|l|}
            \hline
            Hint (\%) & \multicolumn{2}{c|}{\textbf{0\%}} & \multicolumn{2}{c|}{\textbf{25\%}} & \multicolumn{2}{c|}{\textbf{50\%}}& \textbf{70\%}\\
            \hline
            Model & NYT   & \emph{init}   & NYT   & \emph{init}   & NYT   & \emph{init} & \emph{init} \\
            \hline
            Mistral 7B (5-shot) & 10.95\%   & 1.70\%    & 9.70\%    & 2.80\%    & 11.95\%   & 4.80\%    & \\
            LlaMa 3 8B (5-shot) & 15.8\%    & 1.30\%    & 19.7\%    & 2.85\%    & 24.65\%   & 6.25\%    & \\
            LlaMa 3 70B (5-shot) & 27.20\%   & 6.40\%    & 31.80\%   & 11.45\%   & 45.30\%   & 20.35\%   & \\
            GPT 4 Turbo (5-shot) & 41.2\%    & 18.70\%   & 59.95\%   & 33.70\%   & 75.75\%   & 52.85\%   & \textbf{76.30\%}\\
            \hline
            
        \end{tabular}
    \end{center}
    \caption{
        \textbf{Testing if LLMs can improve by using character constraints from 
        partially filled crosswords}. \citet{llmcryptic} reported an accuracy of 
        27.0\% (70\% hinted clues) by fine-tuning a Mistral 7B model on the 
        \emph{Init} dataset, \emph{which GPT-4-Turbo} (\textbf{76.30\%} accuracy) 
        \emph{outperforms by a factor of} \textbf{$\sim$2.8$\times$}
        \emph{without fine-tuning}.
    }
    \label{tab:hinted_results}
\end{table*}

We observe (Table \ref{tab:hinted_results}) that, in all but one case, LLMs show
improved performance with an increasing percentage of constraint information for
both datasets. Additionally, to compare the performance of GPT-4-Turbo to 
previously reported SoTA results by \citet{llmcryptic}, we perform the 
experiment with the same settings and dataset split as them to find that 
\textbf{GPT-4-Turbo} with 5-shot prompts (76.3\% accuracy) outperforms the 
fine-tuned Mistral 7B model (27\% accuracy) by a \textbf{factor of 2.8$\times$}.

The fact that LLMs can successfully exploit constraints to answer crossword 
clues better suggests that they are well-suited to the task of solving full 
crosswords. The significant jump in performance observed in modern SoTA LLMs at 
this task and the task in Section \ref{sec:ansclue} is extremely serendipitous, 
and we investigate this in further detail in Section \ref{sec:contamination}.

\subsection{Sub-token Counting}
\label{sec:subcount}

Despite significant performance gains, SoTA LLMs struggle with adherence to 
length constraints, suggesting an inability to count characters within words or
phrases (sub-token counting). We observed that even the best performing model, 
GPT-4-Turbo, produces answers of incorrect length on 26.2\% of the \emph{Init} 
dataset and 16.9\% of NYT dataset. This may be explained by the tokenization 
methods used in LLMs, such as Byte-Pair Encoding \citep{bpe}. During the word 
embedding stage in transformers \citep{transformer}, tokens are converted into 
embedding vectors, causing the loss of information about individual characters.
This character-level information must be relearned during training. While we
are unsure exactly how LLMs regain this information, we suspect they learn from 
training data that include explicit length details.

There are websites\footnote{\url{https://word.tips/words-by-length/} for 
example.} that contain large lists of words with their corresponding lengths. 
Often replies in message boards also include a count of the number of characters 
in the reply. Artifacts like these, which contain enough information to infer 
the length of tokens, go on to become part of the datasets that LLMs are trained
on. We hypothesize that LLMs learn to count sub-tokens based on this information 
provided during training.

To investigate this further, we devised the sub-token counting task, wherein
the LLM is provided a sequence of (lowercase) characters without whitespaces and 
asked to predict the number of characters making up the sequence. To test our
hypothesis, we consider three sets of 1000 (English) words--\emph{Common}, 
\emph{Medium} and \emph{Rare}--based on word unigram frequencies curated by 
\citet{unigram} from Google's Trillion Words corpus. The \emph{Common}, 
\emph{Medium} and \emph{Rare} words have ranks in the range of 1-5,000,
47,500-52,500, and 95,000-100,000, respectively.

\begin{figure}
    \begin{center}
        \includegraphics[width=0.45\textwidth]{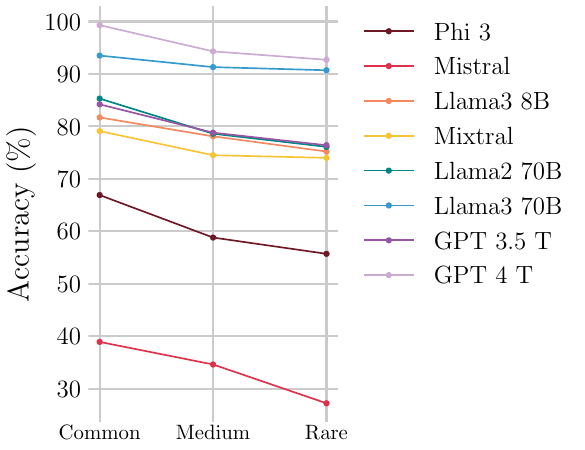}
    \end{center}
    \caption{LLMs ability to count the number of characters in a word declines
    with the frequency of the word.
    }
    \label{fig:count_freq}
\end{figure}

If language models have a widely generalizable ability to perform sub-token 
counting, we should see no difference in counting performance across words with 
different prevalence. However, we observe that (see Figure \ref{fig:count_freq}) 
\emph{the accuracy of LLMs at the sub-token counting task declines with the 
frequency of the token} for all LLMs tested. We further analyzed if there was a 
difference in sub-token counting performance between words that are part of the 
model vocabulary and randomly generated gibberish following the same 
distribution of lengths to account for potential shifts of distribution of 
length between frequent and rare words. To do this, we first created a set of 
words by taking an intersection of all words that are part of the vocabulary of 
every open-source model in consideration and the list of the top 100,000 words. This
is to ensure that the sequences are extremely likely\footnote{We can't be sure
about the proprietary models.} to be vocabulary tokens for every model in 
consideration and are not special tokens like \texttt{<bos>}. Then we created a 
set of gibberish words by replacing each character of the vocab set words with a
randomly chosen character from the set \texttt{\{a-z\}}, thus guaranteeing that
they have the same length distributions.

\begin{table}[ht]
    \begin{center}
        \begin{tabular}[c]{|l|l|l|}
            \hline
            \multirow{2}{*}{\textbf{Model}}                   & \textbf{Vocab.}        & \textbf{Gibberish} \\
                                    & Acc. $(\%)$     & Acc. $(\%)$\\
            \hline
            Phi 3 3.8B Instruct     & $79.4$         & $61.2$ \\
            Mistral 7B Instruct     & $47.9$         & $28.2$ \\
            Llama 3 8B Instruct     & $92.6$         & $69.7$ \\
            Mixtral 8x7B            & $92.6$         & $80.1$ \\
            Llama 2 70B             & $92.8$         & $80.0$ \\
            Llama 3 70B             & $99.6$         & $87.5$ \\
            GPT 3.5 Turbo           & $86.0$         & $62.1$ \\
            GPT 4 Turbo             & $99.8$         & $98.8$ \\
            \hline
        \end{tabular}
    \end{center}
    \caption{LLM sub-token counting performance for vocabulary words and gibberish.}
    \label{tab:count_gibber}
\end{table}

We find (see Table \ref{tab:count_gibber}) that in addition to counting accuracy
being affected by frequency, there is often a large disparity between the accuracy
of \emph{vocabulary} vs. \emph{gibberish} words. Although this does not
conclusively show that LLMs rely on memorized training instances to perform 
sub-token counting, it does provide strong evidence suggesting that LLMs learn
to count based on length information containing artifacts in training data. Future 
works exploring this idea further would be compelling.

\section{SweepClip - Our proposed algorithm to solve crosswords with LLMs}
\label{sec:proposal}

In this section, we address the problem of filling crossword grids with LLM 
assistance. Note that this task not only involves generating correct answers 
from provided clues but also hinges on exploiting constraints from already 
generated words and backtracking to eliminate past generations that do not fit
well when new evidence becomes available. Since LLMs demonstrate this ability,
when paired with the right search algorithm, it should be possible to solve
crosswords with the aid of LLMs.

%\begin{algorithm}
%    \caption{\textbf{ - SweepClip} (Sketch)}
%\label{alg:sweep_short}
%\begin{algorithmic}[1]
%    \State Given crossword grid, clues and an LLM.
%    \State Generate candidate answers for all clues. \emph{(sweep)}
%    \For{$i$ in $\{1, \ldots, \texttt{max\_iter}\}$}
%        \State Prune answers not in largest connected component. \emph{(clip)}
%        \State Prune conflicting answers.
%        \State Generate answers for neighbors of surviving answers.
%        \If{\texttt{solved} or \texttt{budget\_exceeded}}
%            \State \texttt{break}
%		\EndIf
%	\EndFor
%\end{algorithmic}
%\end{algorithm}

Our proposed algorithm (\textbf{detailed in Appendix} \ref{sec:appendix-algo}) 
first generates a set of candidate answers for all clues provided with the 
crossword (\emph{sweep}) and uses a graph-based criterion to eliminate answers 
that do not fit (\emph{clip}). Then we use the constraints generated from the 
previous step answers to generate more candidate answers\footnote{with 
appropriate masks provided as prompts, see Appendix \ref{sec:appendix-algo}.}
and prune the bad-fitting candidates. We iteratively apply this strategy until 
either (i) the entire crossword is filled, (ii) we exceed a preset number of 
iterations, or (iii) we run out of LLM computational budget. 

For pruning, we use the largest connected component from the answers generated
so far to ensure that the generated answers are somewhat coherent amongst
themselves. This is over-restrictive, as it is possible that isolated answers 
are correct; however, we find that this strict pruning strategy works better to
eliminate bad 
answers early rather than using the constraints imposed by potentially bad answers 
to generate further bad answers (see Appendix \ref{sec:appendix-algo}).

\section{Results}
\label{sec:results}

In this section, we present results from our algorithm at the crossword-solving
task and present further results investigating the performance demonstrated by 
current-generation SoTA LLMs.

\subsection{\emph{SweepClip}--Solving NYT Crossword Puzzles}
\label{sec:sweepclipres}

For this task, we employed our algorithm, \emph{SweepClip}, on a set of 100 
randomly sampled Monday \emph{NYT} crossword puzzles. We used two LLMs for this:
GPT-4-Turbo and Llama 3 70B (see Appendix \ref{sec:appendix-algo}).

\begin{table}[ht]
    \begin{center}
        \begin{tabular}[c]{|p{8em}|l|l|}
            \hline
            \textbf{Error Tolerance} & \multicolumn{2}{c|}{\textbf{\% of Crosswords}} \\
            \hline
                                      & LLaMa 3  & GPT-4 T\\
            \hline
            \textbf{100\% solved}     & 0  & \textbf{48}  \\
            \hline
            $\leq$ 1 character error  & 1  & \textbf{55}  \\
            \hline
            $\leq$ 5 character error  & 10 & \textbf{71}  \\
            \hline
            $\geq 90\%$ Accuracy      & 30 & \textbf{80}  \\
            \hline
            $\geq 50\%$ Accuracy      & 82 & \textbf{98}  \\
            \hline
        \end{tabular}
    \end{center}
    \caption{
        Results from solving NYT crosswords with our algorithm \emph{SweepClip}.
    }
    \label{tab:sweep_clip}
\end{table}

We find (see Table \ref{tab:sweep_clip}) that our algorithm with GPT-4-Turbo 
solves \textbf{48\%} of crosswords without any errors and \textbf{55\%} of 
crosswords with at most 1 character wrong. The average character level accuracy 
in crossword solving is \textbf{93.1\%} ($\pm$ 14.1\%). Our algorithm improves
the clue-wise answer accuracy\footnote{clue-level accuracy is different from
character-level accuracy \citep{dacross}; e.g., it is possible to have a 
filled-in crossword without deciphering all clues.} (exact match) to \textbf{89.6\%} 
($\pm$ 16.9\%) from the base accuracy (without the algorithm), which is 43.5\% 
($\pm$ 23.5\%), an improvement of \textbf{2.1}$\times$. Note, the previously 
reported SoTA accuracy on this task with a foundational LM (without fine-tuning)
was \textbf{26\%} with retrieval-augmented generation and an SMT solver coupled
with an oracle that eliminates parts of the crossword grid that do not have
suitable generated answers \citep{dacross}.

The performance for the smaller LLaMA 3 70B is worse; however, our algorithm
still manages to improve final clue answering accuracy to \textbf{59.4\%} 
($\pm$ 24.1\%) from a base accuracy of only 22.3\% ($\pm$ 14.4\%). Note that
this result, alongside those presented in Section \ref{sec:analysis}, serve to
\emph{ablate} our crossword-solving approach.

Thus, with the application of our algorithm, we have successfully exploited the 
constraint information to boost the performance of LLMs beyond what would have 
been possible with straightforward QA like clue deciphering. To the best of our 
knowledge, \textbf{this is the first algorithm that demonstrates successfully
solving crosswords with the aid of an out-of-the-box LLM}.

\subsection{Solving Cryptic Crossword Clues}

\begin{table}[ht]
    \begin{center}
        \begin{tabular}[c]{|p{9em}|p{6em}|l|}
            \hline
            \multirow{2}{*}{\textbf{Model}}       & \multirow{2}{*}{\textbf{Method}}      & \textbf{Acc.}\\
                                 &                      & $(\%)$       \\
            \hline
            Rule-based              & CFG+WordNet       &       \\
            \citep{deits}           &                   & 7.3   \\
            \hline
            T5 \citep{cryptonite} & SFT                 & 1.1   \\
            \hline
            T5                  & Curriculum            &       \\
            \citep{curriculum}  & Learning              & 6.5   \\
            \hline
            Mistral 7B          & SFT                & 1.2   \\
            Mistral 7B          & 10 shot            & 4.6   \\
            Chat GPT            & 3 shot             & 9.5  \\
            \citep{llmcryptic}  &                    &      \\
            \hline
            GPT 4 Turbo \textbf{(ours)}               & 5 shot              & 18.70 \\
            GPT 4 Turbo \textbf{(ours)}               & CoT(1)@3SC    & \textbf{20.85} \\
            \hline
        \end{tabular}
    \end{center}
    \caption{
        \textbf{Comparison of our results with previously reported SoTA results.}
        Results are on the \emph{Init} dataset with crossword clue deciphering 
        treated as QA.
    }
    \label{tab:comp_sota}
\end{table}

We observed (see Table \ref{tab:comp_sota}) that SoTA LLMs have significantly 
improved cryptic crossword clue deciphering abilities. We also note that 
\emph{chain-of-thought} \citep{cot} prompting (1-shot) with self-consistency 
\cite{cotsc} (3 samples) leads to further performance gains. Our best result on 
the \emph{word-init-disjoint} split is \textbf{20.85\%}, which \emph{improves 
over the previous SoTA (9.5\%) by a factor of} \textbf{2.2$\times$} 
\emph{without any fine-tuning}. 

\subsection{Data Contamination and Generalizability}
\label{sec:contamination}

To rule out data contamination as the reason behind the significant performance 
gains by SoTA LLMs, we curated additional cryptic crossword clue datasets 
comprising entirely of puzzles published after \textbf{May 20, 2024, which is 
after the knowledge cutoff date of all LLMs examined}. These datasets are 
sourced from \href{https://www.theguardian.com}{The Guardian} and 
\href{https://www.lovattspuzzles.com}{Lovatts Puzzles}. The answers in these 
\textit{post-cutoff} sets were checked against the combination of all other 
cryptic crossword datasets employed in the study (665,497 answers in total) to 
check for potential duplicates. None was found for the \textit{post-cutoff} 
Guardian set, and 2 were found for the \textit{post-cutoff} Lovatts set, which 
were removed. Note: the \emph{Init} dataset is also sourced from \emph{The 
Guardian}; thus, the results reported in Table \ref{tab:post_cutoff} 
should be consistent\footnote{More details on data curation is in Appendix 
\ref{sec:appendix-data}.}.

\begin{table}[ht]
    \begin{center}
        \begin{tabular}[c]{|l|l|l|l|}
            \hline
            \textbf{Model}           & \emph{Lovatts}    & \emph{Guardian} & \emph{init} \\
            \hline
            Llama 3 70B              &    26.03\%        & 5.5 \%          & 6.4 \%  \\
            Claude 3 Sonnet          &    46.28\%        & 12.5\%          & 10.8\%  \\
            GPT 4 Turbo              &    61.57\%        & 18.5\%          & 18.7\%  \\
            \hline
        \end{tabular}
    \end{center}
    \caption{
        \textbf{Performance (accuracy) of LLMs on curated datasets that appeared 
        after the knowledge cut-off.} Note, \emph{Init} by \citet{curriculum}, 
        also sourced their data from \emph{The Guardian}; thus, these results
        provide a fair head-to-head comparison of performance.
    }
    \label{tab:post_cutoff}
\end{table}

\emph{We see no appreciable difference in performance on the post-cutoff 
dataset} (see Table \ref{tab:post_cutoff}), leading us to suggest that these 
LLMs can generalize beyond potential contamination in their training set. 

\subsection{Human Evaluation and Further Analysis}
\label{sec:heval}

To ascertain if the models can reason about cryptic crossword clues to arrive at
correct responses, we elect to perform \textbf{human evaluation}\footnote{
Further details in Appendix \ref{sec:appendix-hev}} of model responses. We
employ a 3-shot \emph{chain-of-thought} prompt to elicit a reasoned response to
crossword clues from GPT-4-Turbo, which are then analyzed by our team for
soundness vis-\`a-vis factual and logical errors. We choose 100 samples from the
post-cutoff \emph{Lovatts} set and rely on the consensus of all evaluators to
ascertain if a response contains factual or logical errors (e.g., wrong
counting, wrong anagram, drawing a conclusion that does not follow from the
premise, etc.). The author-annotated reasoning responses had low inter-observer
variability (Fleis' $\kappa=0.94$, pre-consensus) and have been made publicly
available.

\begin{table}[ht]
    \begin{center}
        \begin{tabular}[c]{|p{9em}|c|c|}
            \hline
            \multirow{2}{*}{\textbf{GPT 4 Turbo}} & \textbf{Sound}    & $\neg$\textbf{Sound} \\
                            &  (61)             & (39)             \\
            \hline
            Correct (65) &  48\%    & 17\% \\
            \hline
            Wrong (35)   &  13\%    & 22\% \\
            \hline
        \end{tabular}
    \end{center}
    \caption{
        \textbf{Results from human evaluation of reasoning.} 
        \emph{Chain-of-thought} elicited reasoning responses from GPT-4-Turbo
        are evaluated for soundness. An answer is called \textbf{correct} if the
        model prediction exactly matches the ground truth. The answer is called
        \textbf{sound} if it contains no logical or factual errors. Results are
        on the post-cutoff Lovatts' set.
    }
    \label{tab:human-eval}
\end{table}

\begin{table*}[ht]
        \centering
        \begin{tabular}[c]{|l|l|l|l|l|l|}
            \hline
            \multirow{2}{*}{\textbf{GPT-4-Turbo}} & \textbf{ANG}   & \textbf{SCJ}      & \textbf{CNT}  & \textbf{HOM}  & \textbf{OTH}  \\
                                                  & (27)           & (26)              & (18)          & (8)           & (21) \\
            \hline
            & & & & & \\
            EM & 74\%   & 50\%  & 56 \% & 100 \% & 67 \% \\
            & & & & & \\
            \hline
        \end{tabular}
        \hspace{3em}
        \begin{tabular}[c]{|l|l|l|}
            \hline
            \multirow{2}{*}{\textbf{Model}}        &   \textbf{ANG} & \textbf{CNT} \\
                                                   &   (8.5\%)      & (2.5\%)      \\
            \hline
            Llama 3   &   2.6\%        & 12.2\%       \\
            Claude 3  &   7.9\%        & 15.3\%       \\
            GPT 4 T   &   25.0\%       & 33.7\%       \\
            \hline
        \end{tabular}
    \caption{
        \textbf{Do LLMs have common ``failure modes''?} \textbf{ANG} refers to 
        \emph{anagrams}, \textbf{SCJ} refers to \emph{synonym conjugation}, 
        \textbf{CNT} refers to \emph{containment}, \textbf{HOM} refers to 
        \emph{homophones}, and \textbf{OTH} refers to \emph{others}. 
        \emph{(Left)} shows results from GPT-4-Turbo on the post-cutoff Lovatts 
        set, and \emph{(right)} shows results on the combined \emph{Init} and 
        \emph{Cryptonite} datasets. The numbers in parentheses refer to 
        prevalence of the clue type.
    }
    \label{tab:human-eval1}
\end{table*}

The results (see Table \ref{tab:human-eval}) show that \textbf{74\%}
of the time GPT-4-Turbo provided a correct answer, it also gave \textbf{sound reasoning 
in support of the answer}. This leads us to conjecture that they possibly have a
significant ability to reason and generalize.

Furthermore, to analyze if LLMs demonstrate any common failure modes, we manually
annotated the human evaluation dataset based on the principal skill required to 
solve a particular puzzle clue (see Table \ref{tab:human-eval1}). \textbf{ANG}
indicates that the answer is an \emph{anagram} of some words of the clue, 
\textbf{HOM} indicates that the answer is a \emph{homophone} of some words of 
the clue, \textbf{CNT} indicates that the answer is disguised in a contiguous 
section of the clue, and \textbf{SCJ} indicates that the answer is found by 
combining several words that are synonyms of various parts of the clue. The
class \textbf{OTH} lumps together a variety of other kinds of skills (e.g., 
spoonerisms, acronyms, world knowledge, and various other kinds of character 
manipulations). Examples of each are provided in Appendix \ref{sec:appendix-hev}.

Owing to the limited number of human annotations, the results (Table 
\ref{tab:human-eval1} \emph{left}) are not statistically significant, but the
trends suggest that GPT-4-Turbo may exhibit strong performance on \emph{anagrams}
(74\% accuracy vs. 65\% baseline\footnote{Baseline refers to mean accuracy
across all kinds of clues.}) and \emph{homophone}-based clues (100\% accuracy).
We also attempted to perform the same analysis on the \emph{Cryptonite} and 
\emph{Init} datasets (4000 clue-answer pairs) and found similar trends for
anagram performance by GPT-4-Turbo (25\% accuracy vs. 21.1\% baseline). However,
this does not hold for Llama 3 70B (2.6\% accuracy vs. 7.16\% baseline) and 
Claude 3 Sonnet (7.9\% accuracy vs. 11.85\% baseline), suggesting that Llama 3
70B and Claude 3 Sonnet have a diminished ability to deal with anagrams. Note
that unlike \textbf{(ANG, CNT)}, \textbf{(HOM, SCJ, OTH)} cannot be
automatically detected reliably, which limits broader analysis for these kinds
of clues.

To quantify the effect of \emph{sub-token counting performance} on clue-solving
performance, we devise a further experiment. We consider all such clues
for which the model correctly deduced the semantics of the clue but failed to 
adhere to the length constraints (e.g., \texttt{LECTURER} instead of 
\texttt{PROFESSOR} or \texttt{NANNA} instead of \texttt{GRANNY}). We consider 
\emph{wrong} LLM predictions with a semantic similarity of 0.5 or more
\footnote{As given by OpenAI \texttt{text-embedding-3-large} model.} and report
the percentage of answers of incorrect length. GPT-4-Turbo and Llama 3 70B produce
predictions with length errors \textbf{46.4\%} and \textbf{59.9\%} of the time
respectively. This suggests that the lack of adherence to length constraints is 
a major impediment in clue solving for LLMs.

\section{Conclusions}
\label{sec:conclusion}

Solving crosswords requires proficiency in a multitude of desirable skills, and
cryptic crosswords were previously thought to be firmly in the domain of human
experts. Our findings challenge this notion. We have demonstrated that the 
current generation of SoTA LLMs shows significantly improved aptitude at solving 
straight and cryptic crossword puzzle clues and can exploit constraints provided
by partially solved crosswords to boost this performance further. We also found
that this emergent ability generalizes well to the post knowledge cut-off regime
and is accompanied with the capacity to produce reasoned explanations.

We have also developed an algorithm that, with the aid of LLMs, further boosts this
performance and achieves 93\% accuracy on Monday New York Times crosswords. With
the aid of this algorithm, it is possible to achieve significant performance gains
even if the baseline LLM accuracy is low, thus indicating that, when paired
with the right search strategy, LLMs can successfully solve crosswords. Such an
LLM-guided search approach may be readily adapted to other scenarios requiring
certain kinds of constraint satisfaction.

Incorporating length constraints effectively still remains a challenge. We showed
that SoTA LLMs struggle with sub-token counting and also that they often provide 
incorrect predictions despite being semantically close, owing to length constraint 
violations. This weakness also manifests in other tasks like anagrams, character 
manipulations, etc., which are heavily reliant on arithmetic abilities. Future 
research attempting to address this shortcoming would be compelling.

\section*{Limitations}
\label{sec:limit}

\noindent\textbf{Limitation 1 (Algorithm) - } We would like to highlight that our algorithm
is sub-optimal, as it discards potential correct answers and does not fully 
explore the consequences of each possible generated answer. Ideally, we would 
generate one candidate answer, followed by all its neighbors, and whenever there 
is a conflict, branch out and explore all options to see which is a better fit.
We elect not to do this, because it involves a potentially exponential number of
LLM calls. Our algorithm does not provide guarantees of convergence or correctness 
beyond elimination of conflicting answers. Note that this problem is 
\textbf{NP-Hard} and it is difficult to find approximate solutions for it 
\citep{dacross}. This algorithm was designed to minimize computational cost 
incurred in LLM calls; however, future studies with much larger computational 
budget for LLM calls will definitely have greater success employing a more 
thorough search strategy.

\noindent\textbf{Limitation 2 (Cryptic Crossword Solving) - } 
Compared to straight crosswords, the baseline accuracy in clue answering by LLMs
for cryptic crosswords is much lower (e.g., ~18\% for GPT-4-Turbo on cryptics vs
~41\% on straight). Thus, solving cryptic crosswords with current SoTA LLMs would
require a much more extensive search strategy. Our current approach with a budget
limit of 0.5 USD per crossword is unsuccessful (12\% of cryptic crosswords solved
with 50\% letter accuracy) at solving cryptics, and our financial constraints do
not permit a more thorough investigation with improved algorithms at this time.

\noindent\textbf{Limitation 3 (Reporting Crossword Solving Results) - } 
We only reported results on solving Monday New York Times crossword puzzles
and did not report Tuesday-Thursday which typically have a higher level of difficulty.
There are primarily two reasons for this. (i) As we have seen in the paper, 
even if the base LLM has relatively low accuracy, when paired with a search/prune
algorithm, the accuracy can be improved considerably. There is a trade-off between 
base accuracy and the number of LLM calls, i.e., if the base accuracy is lower, 
the algorithm needs to make a lot more calls to the LLM increasing computational
costs. We have observed that it is possible to solve harder NYT (straight) 
crosswords, but at an increased computational budget. This makes reporting 
statistics cost-prohibitive for us at this time. 

\noindent\textbf{Limitation 4 (Sub-token Counting) - } We have demonstrated that 
LLMs show limitations at sub-token counting, and have hypothesised that they do 
not have widely generalizable skill at this task, rather relying on memorization.
The evidence provided in this paper is compelling, but definitely not enough to 
conclusively prove this hypothesis. Further studies are required that can 
intervene at the pre-training stage of LLMs to conclusively demonstrate whether
this hypothesis is true. We do not possess the computational power to undertake 
such a study at this stage.

\noindent\textbf{Limitation 5 (Fine-tuning) - } We do not report results with fine-tuned
models. It is possible that further improvements could be seen with fine-tuning,
however, we are interested in studying the emergent properties of large language 
models in this context instead of building special purpose models that can solve
crosswords.

\section*{Ethics Statement}

In keeping with ACL ethical guidelines we make all scientific artifacts generated
for this study freely available and open source under the MIT licence, this includes 
all software we created, every prompt that was used, all raw model outputs, and 
data we generated. The test sets used for all experiments are also made availabe,
with the exception of the full grids of the New York Times crossword puzzles.
We do not re-distribute the (publicly available) New York Times crossword dataset 
since it is the intellectual property of The New York Times. The use of their 
data falls within the terms of the Fair Use doctrine set within 17 U.S.C. \S 107.
To aid in reproducibility we provide a list of dates to uniquely identify the 
crossword puzzles used in the study, and provide detailed instructions on how 
to acquire them through the New York Times. We do not use any data to train any 
model in this study.

We foresee no serious ethical implications on society at large from this study.

\section*{Acknowledgments}

The authors would like to thank Akshay Chaturvedi and Joy Mahapatra for their 
helpful feedback on this paper. This research was funded in part by the 
Indo-French Centre for the Promotion of Advanced Research (IFCPAR/CEFIPRA) 
through project number \texttt{CSRP 6702-2}.

\bibliography{bibliography/custom}

\appendix

\section{Appendix  - Details of the Sweep Clip Algorithm}
\label{sec:appendix-algo}

This section is intended to provide details of the \emph{SweepClip} algorithm
introduced in Section \ref{sec:proposal}.

A crossword puzzle consists of a grid (see Figure \ref{fig:crossword_subtask})
and a set of clues $C = \{c_1, \ldots, c_n\}$ and answers (ground truth) 
$A = \{a_1, \ldots, a_n\}$. The grid imposes a graph $G = (V, E)$, where 
$V = \{v_1, \ldots, v_n\}$ is the set of vertices corresponding to every 
clue/answer in the crossword, and 
\begin{equation}
    \begin{split}
        E = \{(v_i,v_j)|& \; \forall i,j\;i\neq j\: s.t. \; a_i, a_j \textrm{ share a}\\
        &\textrm{common grid position}\}
    \end{split}
\end{equation}

Given an LLM, and a set of clues $C' \subseteq C, C' =\{c_{j_1}, c_{j_2}, \ldots \}$ 
corresponding to vertices $\{v_{j_1}, v_{j_2}, \ldots \}$, we can generate 
candidate answers $\hat{A} = \{\hat{a}_{j_1}, \hat{a}_{j_2}, \ldots\}$, where 
$\hat{a} = LLM(c)$. We abbreviate this as $\hat{A} = LLM(C')$. For a subset 
$V'$ of $V$, let the set of clues associated with $V$ be denoted by $C(V')$.

In subsequent iterations of the algorithms, when some candidate answers have been
generated, we generate further candidate answers for their neighboring vertices.
In this case, the partially unmasked characters are provided to the LLM as in 
Section \ref{sec:partgrid}. Note, our algorithm implicitly performs 
self-consistency checks to improve candidate answers, e.g., an answer generated 
at the first sweep may be discarded only to be accepted in later iterations, 
when it is coherent with a larger number of other answers.

Two candidate answers $\hat{a}_i, \hat{a}_j$ are said to be in \textbf{conflict},
if $\mu$-th position of $a_i$ and $\nu$-th position of $a_j$ are in the same grid
position, however $\hat{a}_i [ \mu ] \neq \hat{a}_j [ \nu ]$, i.e. the $\mu$-th
character of $a_i$ and $\nu$-th character of $a_j$ are different.

There are two sub-graphs $G_p(\hat{A})$ and $G_n(\hat{A})$ of $G$ that correspond
to a set of candidate answers $\hat{A}$. Let $v_i,v_j$ correspond to $\hat{a}_i,
\hat{a}_j \in \hat{A}$. An edge $(v_i, v_j) \in E$ is in $G_p(\hat{A})$
if and only if $\hat{a}_i, \hat{a}_j$ don't conflict, else it is in $G_n(\hat{A})$.
We call the largest connected component of a graph $H$, $LCC(H)$, and for a 
subset $S$ of vertices $V$ of a graph $G$, $ngbd(G, S)$ denotes the vertices in
$V$ that are adjacent to $S$ but not in $S$. The algorithm is detailed in 
Algorithm \ref{alg:sweep_clip}.

\begin{algorithm}
    \caption{\textbf{- SweepClip} (Detailed)}
\label{alg:sweep_clip}
\begin{algorithmic}[1]
    \State Given $C$, crossword graph $G$ and an LLM.
    \State Generate $\hat{A} \leftarrow LLM(C)$.
    \For{$i$ in $\{1, \ldots, \texttt{max\_iter}\}$}
        \State Construct $G_p(\hat{A}), G_n(\hat{A})$
        \State $L \leftarrow LCC(G_p(\hat{A}))$.
        \State $\hat{A} \leftarrow \{\hat{a}_i |\;v_i \in L\}$
        \While{$G_n(\hat{A})$ has edges.}
            \State A max degree vertex in $G_n(\hat{A}) \rightarrow v_m$
            \State Remove $v_m$ from $G_n(\hat{A})$.
            \State $\hat{A} \leftarrow \hat{A} - \hat{a}_m$
        \EndWhile
        \State Calculate $N \leftarrow ngbd(G, \hat{A})$.
        \State Add character information to $C(N)$.
        \State $\hat{A} \leftarrow \hat{A} \cup LLM(C(N))$
        \If{\texttt{solved} or \texttt{budget\_exceeded}}
            \State \texttt{break}
		\EndIf
	\EndFor
\end{algorithmic}
\end{algorithm}

The results in Section \ref{sec:sweepclipres} are produced with \texttt{max\_iter}
of 30 and a budget of 0.5 USD per crossword for GPT-4-Turbo, and \texttt{max\_iter}
of 35 and a budget of 600 LLM calls for LLaMA 3 70B.

\section{Appendix  - Datasets}
\label{sec:appendix-data}

\begin{table}[h]
    \begin{center}
        \begin{tabular}[c]{|p{6em}|l|l|l|}
            \hline
            \textbf{Dataset}    & \multirow{2}{*}{\textbf{Train}}     & \multirow{2}{*}{\textbf{Validation}} & \multirow{2}{*}{\textbf{Test}}   \\
            (Clues)    & & & \\
            \hline
            Cryptonite         & 470,804          & 26,156        & 26,157 \\
            \hline
            word-init-disjoint & 75,847           & 32,628        & 33,905 \\
            \hline
            NYT (Clues)        & \multicolumn{2}{|l|}{10,000}     & 2,000  \\
            \hline
            NYT (Grids)        & \multicolumn{3}{|l|}{100 (7700 clues) \emph{Test only}} \\
            \hline
            \multicolumn{4}{|c|}{\textbf{After May 20, 2024}}\\
            \hline
            Lovatts             &                 &               & 242  \\
            \hline
            The Guardian        &                 &               & 200  \\
            \hline
        \end{tabular}
    \caption{Dataset details.}
    \label{tab:dataset_details}
    \end{center}
\end{table}

This section contains details of the datasets used, and details relating to how
they were curated.

\subsection{Cryptonite and word-init-disjoint}

\emph{Cryptonite}\footnote{\url{https://github.com/aviaefrat/cryptonite}} is a 
cryptic crossword clue dataset introduced by \citet{cryptonite}, and is gathered
from \emph{The Times} and \emph{The Telegraph} - two popular UK dailies. In 
their official split, the training and test sets do not contain any common 
answers, so that a model trained on the \texttt{train} split cannot rely on 
memorization to answer the cryptic crossword clue. Two thousand random samples 
were chosen from the \texttt{train} split to report results in Section 
\ref{sec:analysis}.

\emph{word-init-disjoint (Init)} is also a cryptic crossword clue dataset and 
was introduced by \citet{curriculum}. This is gathered from \emph{The Guardian},
and their official split introduces the additional constraint that if two 
answers share a common root (first two characters) they are placed in the same 
split. The results reported in \ref{sec:analysis} are based on two thousand 
randomly sampled clues from their official \texttt{test} split. We distribute 
all materials pertaining to this dataset\footnote{\url{https://github.com/jsrozner/decrypt}}
with this paper. 

\subsection{New York Times Dataset}

We collected and curated two datasets from the \emph{New York Times}
\footnote{\url{https://www.nytimes.com/crosswords/}} for our
analysis of straight crossword puzzles. The first consists of 100 randomly sampled
Monday crossword puzzles ranging from 20th January 1969 to 7th August 2023 with
all clues (7700) and grid information, which were used to report results in Section 
\ref{sec:results}.

Additionally, we curated 12,000 randomly sampled clue answer pairs split into 
two sets - \texttt{test} (2000) and \texttt{support} (10,000). All three sets 
are completely disjoint.\footnote{\url{https://www.xwordinfo.com/} is useful 
for older NYT crosswords}

The \texttt{test} and \texttt{support} sets are distributed alongside this paper
in the github repository, however, the 100 complete grids are not, as they
are intellectual property of the New York Times. They can be accessed after 
purchasing a subscription to their service. To aid reproducibility we include the
list of dates to uniquely identify the crossword grids used to report results.

\subsection{Post-cutoff datasets}

We curated two post-cutoff datasets to test for potential dataset contamination.
One of them was sourced from Lovatts puzzles \footnote{\url{https://lovattspuzzles.com/}}
and \emph{The Daily Mail}. In total, 244 clues and answers were manually curated
from puzzles published strictly after May 20, 2024. To maintain parity with 
the \emph{Init} dataset by \citet{curriculum} we also obtained 200 clue-answer 
pairs from \emph{The Guardian} published after May 20, 2024.

We cross-checked both these datasets for duplicate clues against all 665,497 clues
from the \emph{Cryptonite} and \emph{Init} datasets. We found two duplicates for 
the \emph{Lovatts} set which were removed. The dataset obtained from 
\emph{The Guardian} contained no duplicates. This does not guarantee that 
these cryptic crossword clues have not appeared in any form previously, but we
estimate that probability to be low, since there were very few duplicates when 
matched against the extremely large cryptic crossword datasets. Both these 
datasets are distributed alongside the paper in the github repository.

\subsection{Sub-token counting}

For the task in Section \ref{sec:subcount} we first obtained a list of 300,000+
words and their unigram frequencies provided by \citet{unigram} and sourced from
the Google Trillion Words corpus. For the \emph{Rare}, \emph{Medium} and 
\emph{Common} words we randomly selected 1000 words each from ranks 95,000 - 
100,000, 47,500 - 52,500 and 0 - 5,000, respectively. The \emph{Vocab} set was 
constructed by taking an intersection of the vocabulary tokens of every open 
source language model in consideration, i.e., Phi 3, LLaMA 3 70B, LLaMA 3 8B, 
LLaMA 2, Mistral 7B, and Mixtral. Further, we took the intersection of this set 
with the set of top 100,000 words to make sure the chosen tokens are actual 
words and not special tokens. We sampled 1000 words from the resulting
set to create the \emph{vocab} set. For the \emph{gibberish} set, we replaced 
each character from each word in the \emph{vocab} set with a character randomly 
sampled from the set \texttt{\{a-z\}}. This ensures that the distribution of 
lengths of words seen in the \emph{gibberish} set is the same as the 
distribution of lengths in the \emph{vocab} sets. These 5 sets are distributed
alongside the paper in the github repository.

\section{Appendix  - Experimental Details}
\label{sec:appendix-exp}

\subsection{General Overview}

\begin{table}[h]
    \begin{center}
        \footnotesize{
        \begin{tabular}[c]{|p{5.5em}|l|l|p{4em}|}
            \hline
            Model   & Params.                   & Context        & Knowledge       \\
                    &                           & Length         & Cut-off         \\
            \hline
            Phi 3 mini Instruct                 & 3.8B  & 4K    & Oct. 2023     \\
            \hline
            Mistral v0.2 Instruct               & 7B    & 32K   & Dec. 2023    \\
            \hline
            LLaMA 3 {Instruct}                  & 8B    &  8K   & Mar. 2023       \\
            \hline
            Mixtral v0.1                        & 8x7B  &  32K  & Dec. 2023    \\
            \hline
            LLaMA 2                             & 70B   &   4K  & Sep. 2022   \\
            \hline
            LLaMA 3                             & 70B   &   8K  & Dec. 2023    \\
            \hline
            Claude 3 sonnet \texttt{20140229}   &   ?   & 200K  & Mar. 2024       \\
            \hline
            GPT-3.5-Turbo-\texttt{0125}       &   ?   &  16K  & Sep. 2021   \\
            \hline
            GPT-4-Turbo \texttt{2024-04-09}     &   ?   &  128K & May 2024         \\
            \hline
        \end{tabular}}
    \caption{Models used and their details.}
    \label{tab:models}
    \end{center}
\end{table}

We used \texttt{temperature=0.5} for all models and experiments. For few-shot 
prompt responses, \texttt{max-tokens} were set to 10, and for chain-of-thought 
responses, \texttt{max-tokens} was set to 1000. All other paramters including 
\texttt{top\_p, top\_k}, etc., were set to their defaults. The \texttt{Jinja} 
templates used to format the text prompts are provided alongside the paper, and 
are borrowed from the model-cards of the respective models. Claude and GPT 
models were accessed through their respective APIs and all other models were run 
locally on one server consisting of 2x80GB Nvidia A100 GPUs. All models were 
used in \texttt{bf16} format whenever supported. Format of prompts are given in 
Appendix \ref{sec:appendix-prompts}.

\subsection{Experimental Protocols for Section \ref{sec:partgrid}}

When creating a query for a particular test instance with $k\%$ hints, we 
randomly selected $N$ few-shot instances and ensured that few shot examples also 
had $k\%$ hints. The number of characters revealed $(h)$ is given by the 
formula - 

\begin{equation}
\begin{split}
    &h = \\
    &\max\Bigg(1, \text{round}\Big(\frac{k}{100}\times len(\text{answer})\Big)\Bigg)\\
    & \forall k>0
\end{split}
\end{equation}

$h$ many characters are randomly selected and revealed, all other characters are
replaced with ``\texttt{\_}''.

\section{Appendix  - Prompts}
\label{sec:appendix-prompts}

In this section we will detail the prompts used for all experiments presented 
in the paper. Every prompt used in generating outputs presented in this paper
is provided alongside the paper in the github repository.

\subsection{Few Shot Prompts}

For the experiments presented in Section \ref{sec:ansclue}, we used prompts in 
the following format:

            \begin{verbatim}
[{
    "role":"system", "content":
    "You are an expert crossword 
    solver. Given a clue please 
    provide the best possible 
    answer succinctly. Do not 
    produce extra text.\n The 
    number of characters in the 
    answer is given in brackets 
    and must be strictly adhered 
    to. e.g. Clue: Daily update 
    (4)// means the answer should 
    have 4 characters."
}
{
   "role":"user", "content":
    "Clue: <clue 1> (length) // answer 1\n
    Clue: < ...  > (length) // answer k\n
    Clue: <query clue> (length) // "
}]
            \end{verbatim}

For the experiments presented in Section \ref{sec:partgrid}, we used prompts in 
the following format (with the same system prompt):

            \begin{verbatim}
[{
"role":"user", "content":
"Clue: <clue 1> (3) // _ _ N
        => MEN\n
Clue: < ...  > (6) //
        _ _ E _ _ _ => BREATH\n
Clue: <query clue> (length) //
            _ _ X _ X => "
}]
            \end{verbatim}

\subsection{Chain-of-Thought Prompts}

For the results presented in Table \ref{tab:comp_sota} and for human-evaluation
we used \emph{chain-of-thought} prompts. These were based on 10 examples we 
manually created and are diverse in terms of skills required to solve them. An
example \emph{chain-of-thought} reasoning prompt is as follows:
            \begin{verbatim}
"Clue: ecstasy in old forbidden city,
looking where one shouldn't (7) //\n\n
Let's think step by step.\n
Typically a cryptic crossword 
has synonyms and wordplay 
indicators.\n'ecstasy' can mean 
pleasure, bliss, elation.\n
'forbidden city' is a part 
of Beijing, China.\n'ecstasy'
can also mean the drug MDMA 
or commonly known as e.\n
'ecstasy in old forbidden city'
can mean e in Beijing.\n
Beijing is also called Peking.\n
e in Peking can be peeking 
which is a synonym of 
'looking where one shouldn't '.\n
peeking is 7 characters and 
therefore fits the character 
count.\nAnswer: PEEKING"
            \end{verbatim}

\section{Appendix  - Human Evaluation and Reasoning Examples}
\label{sec:appendix-hev}

Consider the following example cryptic crossword clue - \texttt{``Culminating 
point of story about Judy's husband by the railway track (5,4)''}. 
\footnote{Example taken from \url{https://lovattspuzzles.com}.} Solving this 
clue requires connecting \texttt{`Judy'} to the popular puppet show \emph{Punch 
and Judy}, and inferring that \texttt{`Judy's husband'} refers to \texttt{PUNCH}.
Additionally, we must observe that \texttt{`railway track'} is synonymous to 
\texttt{LINE}, and combining these gives \texttt{PUNCHLINE} which also means 
\texttt{`Culminating point of story'}. In this section we provide a few Chain-of-thought
elicited reasoning examples given by GPT-4-Turbo on the Lovatts dataset. These 
are sampled from the set used to perform human evaluation. 

Further, these samples are manually tagged with the principal skill required to 
solve them. The skill-based categories are:
\begin{itemize}
    \item \textbf{ANG} - the answer is an \emph{anagram} of some words of the
clue (e.g., \texttt{CUBIT IS mixed up cookie } $\rightarrow$ \texttt{BISCUIT}).
    \item \textbf{HOM} - the answer is a \emph{homophone} of some words of the 
clue (e.g., \texttt{Heard PRINTS are for royalty} $\rightarrow$ \texttt{PRINCE}).
    \item \textbf{CNT} - the answer is disguised in a contiguous section of the 
clue (e.g., \texttt{The Press leaves presENTERs to go in} $\rightarrow$ \texttt{ENTER}).
    \item \textbf{SCJ} - the answer is found by combining several words which are
synonyms of various parts of the clue (e.g., \texttt{Reasonable FOOD ALLOWANCE$_1$ for CAPONE$_2$} $\rightarrow$ \texttt{RATION$_1$AL$_2$}).
    \item \textbf{OTH} - this class lumps together variety of other kinds of 
skills like Spoonerisms, acronyms, world knowledge and various other kinds of character manipulations
(e.g. \texttt{PIpER loses heart on jetty} $\rightarrow$ \texttt{PIER}).
\end{itemize}

Human evaluation was performed by 3 of the authors of this paper. They were 
asked if a particular response contained logical or factual errors and provided 
detailed instructions and examples of the kinds of errors to be wary of. All 
responses were evaluated by all authors and in case of conflicting answers 
(4 out of 100, Fleiss' $\kappa$ = 0.94), discussions were held to reach a 
consensus. All evaluated responses alongside human-annotated labels are provided 
alongside this paper in the github repository.

In checking whether a justification for an answer given by the LLM is logically 
and factually sound, we assess grammatical soundness and phraseological 
meaningfulness of the sentences in the answer, existence of counterfactual 
statements (e.g., “BULKY has 4 characters”, “the initial letters of ARE RATS 
TIRED NOW are ARTS” etc.), presenting a statement as an inference from previous 
statements when it does not follow from those, etc. If an answer by the LLM is 
found unsatisfactory in any of these aforementioned areas, it is labeled 
\texttt{FALSE}. Following are a few samples of the same.

\par\noindent
\\
\textbf{1. Clue: }%
Initially, any region extending around land%
\\%
\textbf{Answer (Ground Truth): }%
AREA%
\\%
\textbf{Human Consensus (Soundness): }%
FALSE%
\\\\%
\textbf{GPT{-}4{-}Turbo Output: }%
\\\\%
\texttt{'Initially' is often used in cryptic crosswords to indicate taking the initial letters of the following words.}%
\\%
\texttt{'any region extending around land' gives us the words Any, Region, Extending, and Land.}%
\\%
\texttt{\color{red}{Taking the initial letters of each word, we get A, R, E, and A.}}%
\\%
\texttt{Putting these letters together forms the word AREA.}%
\\%
\texttt{'Area' is a synonym for 'region' and fits the clue both in definition and letter count.}%
\\%
\texttt{AREA is 4 characters and therefore fits the character count.}%
\\%
\texttt{Answer: AREA}%
\\%

\par\noindent
\\
\textbf{2. Clue: }%
Fur stayed in audition for medical help%
\\%
\textbf{Answer (Ground Truth): }%
FIRSTAID%
\\%
\textbf{Human Consensus (Soundness): }%
FALSE%
\\\\%
\textbf{GPT{-}4{-}Turbo Output: }%
\\\\%
\texttt{'Fur' can mean pelt, coat, hide.}%
\\%
\texttt{'stayed' can mean remained, lingered.}%
\\%
\texttt{'in audition' indicates a homophone clue (sounds like).}%
\\%
\texttt{'for medical help' can mean aid, care, assistance.}%
\\%
\texttt{Combining these:}%
\\%
\texttt{{-} 'Fur stayed in audition' suggests a homophone of a phrase where 'fur' and 'stayed' combine to sound like another word.}%
\\%
\texttt{{-} 'Fur' as 'fur' and \color{red}{'stayed' sounds like 'aid' (homophone of stayed)}.}%
\\%
\texttt{{-} Putting them together doesn't form a clear word, so reconsidering 'fur' might lead to another homophone, such as 'first aid' \color{red}{(fur sounded + stayed sounding} like first aid).}%
\\%
\texttt{}%
\\%
\texttt{'First aid' fits as a term for 'medical help'.}%
\\%
\texttt{'First aid' is 8 characters and therefore fits the character count.}%
\\%
\texttt{Answer: FIRST AID}%
\\%
\par\noindent
\\
\textbf{3. Clue: }%
Be earlier because of self{-}esteem%
\\%
\textbf{Answer (Ground Truth): }%
FOREGO%
\\%
\textbf{Human Consensus (Soundness): }%
TRUE%
\\\\%
\textbf{GPT{-}4{-}Turbo Output: }%
\\\\%
\texttt{'Be earlier' can suggest synonyms such as precede, antedate.}%
\\%
\texttt{'because of' is often used in cryptic clues as a linking phrase or to indicate causation.}%
\\%
\texttt{'self{-}esteem' can be linked to synonyms like pride, ego.}%
\\%
\texttt{}%
\\%
\texttt{Breaking down the phrase:}%
\\%
\texttt{'Be earlier' suggests a word meaning to come before {-} "precede" fits but is too long.}%
\\%
\texttt{Let's consider "ego" for 'self{-}esteem', which is a common synonym.}%
\\%
\texttt{Looking at 'Be earlier because of self{-}esteem', we can interpret this as a word that indicates being earlier due to ego, or ego causing one to be first or ahead.}%
\\%
\texttt{The answer "PRE" (as in prefix, or before) combined with "EGO" gives us PREGO, which is 5 letters and does not fit. However, considering synonyms and length, the correct answer could be "PRECEDE" if we think of it as "PRE" (before) + "CEDE" (yield, coming from ego or self{-}esteem yielding to others). But this is 7 letters.}%
\\%
\texttt{Reevaluating, if we simplify to just consider "ego" leading or being earlier, we can think of "PREGO" as a playful or informal way to say "before" due to ego, but the character count is off.}%
\\%
\texttt{The best fit for a 6{-}letter word considering the clue and synonyms would be "BEFORE," which directly means "be earlier" and can be influenced by one's self{-}esteem (ego) driving their actions to be ahead or first. This interpretation fits the clue structure and the character count.}%
\\%
\texttt{Answer: BEFORE}%
\\%

\section{Appendix  - Results}
\label{sec:appendix-results}

\begin{table}[ht]
    \begin{center}
        \begin{tabular}[c]{|l|l|l|l|}
\hline
Model                 & Common & Medium & Rare   \\
\hline
Phi 3                 & 66.9\%  & 58.8\%  & 55.7\%  \\
Mistral 7B            & 38.9\%  & 34.6\%  & 27.2\%  \\
Llama 3 8B            & 81.7\%  & 78.1\%  & 75.2\%  \\
Mixtral 8x7B          & 79.1\%  & 74.5\%  & 74.0\%  \\
Llama 2 70B           & 85.3\%  & 78.6\%  & 76.1\%  \\
Llama 3 70B           & 93.5\%  & 91.3\%  & 90.7\%  \\
GPT 3.5 Turbo         & 84.2\%  & 78.8\%  & 76.4\%  \\
GPT 4 Turbo           & 99.3\%  & 94.3\%  & 92.7\%  \\
\hline
        \end{tabular}
    \end{center}
    \caption{Data for results plotted in Figure \ref{fig:count_freq}.}
    \label{tab:cou_results}
\end{table}

\begin{table*}[b]
    \begin{center}
        \begin{tabular}[c]{|l|l|l|l|l|}
            \hline
Model                 & Method          & Accuracy (NYT) & Accuracy (Crypt)  & Accuracy (Init) \\ 
                      &                 & (\%)           & (\%)              & (\%)            \\ 
            \hline
Phi 3 3.8B Instruct   & 5 shot          & 5.15          & 0.80             & 0.35           \\ 
                      & 10 shot         & 5.55          & 0.90             & 0.30           \\ 
            \hline
Mistral 7B Instruct   & 5 shot          & 11.50         & 1.65             & 1.70           \\ 
                      & 10 shot         & 11.60         & 2.05             & 2.00           \\ 
            \hline
Llama 3 8B Instruct   & 5 shot          & 15.90         & 1.80             & 1.30           \\ 
                      & 10 shot         & 14.45         & 2.20             & 1.30           \\ 
            \hline
Mixtral 8x7B          & 5 shot          & 12.70         & 3.85             & 2.65           \\ 
                      & 10 shot         & 13.30         & 4.65             & 2.10           \\ 
            \hline
Llama2 70B            & 5 shot          & 17.75         & 2.95             & 2.85           \\ 
                      & 10 shot         & 18.40         & 3.95             & 2.35           \\ 
            \hline
Llama3 70B            & 5 shot          & 27.20         & 7.85             & 6.40           \\ 
                      & 10 shot         & 28.20         & 8.50             & 5.85           \\ 
            \hline
GPT 3.5 Turbo         & 5 shot          & 26.05         & 7.90             & 5.55           \\ 
                      & 10 shot         & 26.65         & 7.95             & 6.40           \\ 
            \hline
Claude 3 Sonnet       & 5 shot          & 37.70         & 12.9             & 10.8           \\ 
            \hline
GPT 4 Turbo           & 5 shot          & 41.20         & 23.5             & 18.7           \\ 
            \hline
        \end{tabular}
    \end{center}
    \caption{Data for results plotted in Figure \ref{fig:semantic_context}.}
    \label{tab:sem_results}
\end{table*}

\end{document}